\documentstyle[eaclap]{article}
\author{David Carter \\
SRI International Cambridge Computer Science Research Centre \\
23 Millers Yard, Mill Lane \\
Cambridge CB2 1RQ, U.K. \\
{\tt dmc@cam.sri.com}}
\title{Rapid Development of Morphological Descriptions for \\
Full Language Processing Systems}

\begin{document}

\maketitle
\bibliographystyle{acl}

\begin{abstract}

I describe a compiler and development environment for
feature-augmented two-level morphology rules integrated into a full
NLP system. The compiler is optimized for a class of languages
including many or most European ones, and for rapid development and
debugging of descriptions of new languages.  The key design decision
is to compose morphophonological and morphosyntactic information, but
not the lexicon, when compiling the description.  This results in
typical compilation times of about a minute, and has allowed a
reasonably full, feature-based description of French inflectional
morphology to be developed in about a month by a linguist new to the
system.

\end{abstract}

\section{Introduction}
\label{intro}

The paradigm of two-level morphology (Koskenniemi, 1983) has become
popular for handling word formation phenomena in a variety of
languages. The original formulation has been extended to allow
morphotactic constraints to be expressed by feature specification
(Trost, 1990; Alshawi {\it et al}, 1991) rather than Koskenniemi's less
perspicuous device of continuation classes. Methods for the automatic
compilation of rules from a notation convenient for the rule-writer
into finite-state automata have also been developed, allowing the
efficient analysis and synthesis of word forms. The automata may be
derived from the rules alone (Trost, 1990), or involve composition
with the lexicon (Karttunen, Kaplan and Zaenen, 1992).

However, there is often a trade-off between run-time efficiency and
factors important for rapid and accurate system development, such as
perspicuity of notation, ease of debugging, speed of compilation and
the size of its output, and the independence of the morphological and
lexical components. In compilation, one may compose any or all of
\begin{itemize}
\item[(a)] the two-level rule set,
\item[(b)] the set of affixes and their allowed combinations, and
\item[(c)] the lexicon;
\end{itemize}
see Kaplan and Kay (1994) for an exposition of the mathematical basis.
The type of compilation appropriate for rapid development and
acceptable run-time performance depends on, at least, the nature of
the language being described and the number of base forms in the
lexicon; that is, on the position in the three-dimensional space
defined by (a), (b) and (c).

For example, English inflectional morphology is relatively simple;
dimensions (a) and (b) are fairly small, so if (c), the lexicon, is
known in advance and is of manageable size, then the entire task of
morphological analysis can be carried out at compile time, producing a
list of analysed word forms which need only be looked up at run time,
or a network which can be traversed very simply.  Alternatively, there
may be no need to provide as powerful a mechanism as two-level
morphology at all; a simpler device such as affix stripping (Alshawi,
1992, p119ff) or merely listing all inflected forms explicitly may be
preferable.

For agglutinative languages such as Korean, Finnish and Turkish (Kwon
and Karttunen, 1994; Koskenniemi, 1983; Oflazer, 1993), dimension (b)
is very large, so creating an exhaustive word list is out of the
question unless the lexicon is trivial. Compilation to a network may
still make sense, however, and because these languages tend to exhibit
few non-concatenative morphophonological phenomena other than vowel
harmony, the continuation class mechanism may suffice to describe the
allowed affix sequences at the surface level.

Many European languages are of the inflecting type, and occupy still
another region of the space of difficulty. They are too complex
morphologically to yield easily to the simpler techniques that can
work for English.  The phonological or orthographic changes involved
in affixation may be quite complex, so dimension (a) can be large, and
a feature mechanism may be needed to handle such varied but
interrelated morphosyntactic phenomena such as umlaut (Trost, 1991),
case, number, gender, and different morphological paradigms. On the
other hand, while there may be many different affixes, their
possibilities for combination within a word are fairly limited, so
dimension (b) is quite manageable.

This paper describes a representation and associated compiler intended
for two-level morphological descriptions of the written forms of
inflecting languages. The system described is a component
of the Core Language Engine (CLE; Alshawi, 1992), a general-purpose
language analyser and generator implemented in Prolog which supports
both a built-in lexicon and access to large external lexical
databases.  In this context, highly efficient word analysis and
generation at run-time are less important than ensuring that the
morphology mechanism is expressive, is easy to debug, and allows
relatively quick compilation. Morphology also needs to
be well integrated with other processing levels. In particular, it
should be possible to specify relations among morphosyntactic and
morphophonological rules and lexical entries; for the convenience of
developers, this is done by means of feature equations.  Further, it
cannot be assumed that the lexicon has been fully specified when the
morphology rules are compiled.  Developers may wish to add and test
further lexical entries without frequently recompiling the rules, and
it may also be necessary to deal with unknown words at run time, for
example by querying a large external lexical database or attempting
spelling correction (Alshawi, 1992, pp124-7).  Also, both analysis and
generation of word forms are required. Run-time speed need only be
enough to make the time spent on morphology small compared to
sentential and contextual processing.

These parameters -- languages with a complex morphology/syntax
interface but a limited number of affix combinations, tasks where the
lexicon is not necessarily known at compile time, bidirectional
processing, and the need to ease development rather than optimize
run-time efficiency -- dictate the design of the morphology compiler
described in this paper, in which spelling rules and possible affix
combinations (items (a) and (b)), but not the lexicon (item (c)), are
composed in the compilation phase.  Descriptions of French, Polish and
English inflectional morphology have been developed for it, and I show
how various aspects of the mechanism allow phenomena in these
languages to be handled.

\section{The Description Language}

\subsection{Morphophonology}
\label{spellrules}

The formalism for {\it spelling rules} (dimension (a)) is a syntactic
variant of that of Ruessink (1989) and Pulman (1991). A rule is of the
form
\pagebreak
\begin{quote}{\tt
spell($Name$, $Surface$ $Op$ $Lexical$, \\
\hspace*{0.5in} $Classes$, $Features$).
}\end{quote}
Rules may be optional ($Op$ is ``$\Rightarrow$'') or obligatory
($Op$ is ``$\Leftrightarrow$'').  $Surface$ and $Lexical$ are both strings
of the form
\begin{center}
\verb!"!$LContext$\verb!|!$Target$\verb!|!$RContext$\verb!"!
\end{center}
meaning that the surface and lexical targets may correspond if the
left and right contexts and the $Features$ specification are
satisfied. The vertical bars simply separate the parts of the string
and do not themselves match letters. The correspondence between
surface and lexical strings for an entire word is licensed if there is
a partitioning of both so that each partition (pair of corresponding
surface and lexical targets) is licensed by a rule, and no partition
breaks an obligatory rule. A partition breaks an obligatory rule if
the surface target does not match but everything else, including the
feature specification, does.

The $Features$ in a rule is a list of $Feature=Value$ equations.  The
allowed (finite) set of values of each feature must be prespecified.
$Value$ may be atomic or it may be a boolean expression.

Members of the surface and lexical strings may be characters or
classes of single characters. The latter are represented by a single
digit $N$ in the string and an item $N$\verb!/!$ClassName$ in the
$Classes$ list; multiple occurrences of the same $N$ in a single rule
must all match the same character in a given application.

Figure \ref{rules} shows three of the French spelling rules
developed for this system. The {\tt change\_e\_\`{e}1} rule
(simplified slightly here) makes it obligatory for a lexical {\it e}
to be realised as a surface {\it \`{e}} when followed by $t$, $r$, or
$l$, then a morpheme boundary, then $e$, as long as the feature
\verb!cdouble!  has an appropriate value.
The {\tt default} rule that copies characters between surface and lexical
levels and the {\tt boundary} rule that deletes boundary markers are both
optional.
\begin{figure*}
\hspace*{1in}{\tt spell(change\_e\_\`{e}1, "|\`{e}|" $\Leftrightarrow$
"|e|1+e",
[1/trl], [cdouble=n]).} \\
\hspace*{1in}{\tt spell(default, "|1|" $\Rightarrow$ "|1|", [1/letter], []).}
\\
\hspace*{1in}{\tt spell(boundary, "||" $\Rightarrow$ "|1|", [1/bmarker], []).}
\caption{Three spelling rules}
\label{rules}
\end{figure*}
Together these rules permit the following realization of {\it cher}
(``expensive'') followed by {\it e} (feminine gender suffix) as {\it
ch\`{e}re}, as shown in Figure \ref{chere1}.
\begin{figure*}
\begin{center}
\begin{tabular}{|l|c|c|c|c|c|c|c|}\hline
Surface: & c & h & \`{e} & r & & e & \\
Lexical: & c & h &  e    & r & + & e & + \\ \hline
Rule:    & $def.$ & $def.$ & {\it c.e\_\`{e}1} & $def.$ & $bdy.$ & $def.$ &
$bdy.$
\\ \hline
\end{tabular}
\caption{Partitioning of {\it ch\`{e}re} as {\tt cher+e+}}
\label{chere1}
\end{center}
\end{figure*}
Because of the obligatory nature of {\tt change\_e\_\`{e}1}, and the
fact that the orthographic feature restriction on the root {\it cher},
\verb![cdouble=n]!, is consistent with the one on that rule,
an alternative realisation {\it chere}, involving the use of the {\tt
default} rule in third position, is ruled out.\footnote{The {\tt cdouble}
feature is in fact used to specify the spelling changes when {\it e} is added
to various stems: {\tt cher+e}={\it ch\`ere}, {\tt achet+e}={\it
ach\`ete}, but {\tt jet+e}={\it jette}.}

Unlike many other flavours of two-level morphology, the $Target$ parts
of a rule need not consist of a single character (or class
occurrence); they can contain more than one, and the surface target
may be empty. This obviates the need for ``null'' characters at the
surface. However, although surface targets of any length can usefully
be specified, it is in practice a good strategy always to make lexical
targets exactly one character long, because, by definition, an
obligatory rule cannot block the application of another rule if their
lexical targets are of different lengths. The example in Section
\ref{eauelle} below clarifies this point.

\subsection{Word Formation and Interfacing to Syntax}
\label{morphrules}

The allowed sequences of morphemes, and the syntactic and semantic
properties of morphemes and of the words derived by combining them,
are specified by morphosyntactic {\it production rules} (dimension
(b)) and lexical entries both for affixes (dimension (b)) and for
roots (dimension (c)), essentially as described by Alshawi (1992)
(where the production rules are referred to as ``morphology rules'').
Affixes may appear explicitly in production rules or, like roots, they
may be assigned complex feature-valued categories. Information,
including the creation of logical forms, is passed between
constituents in a rule by the sharing of variables. These
feature-augmented production rules are just the same device as those
used in the CLE's syntactico-semantic descriptions, and are a much
more natural way to express morphotactic information than finite-state
devices such as continuation classes (see Trost and Matiasek, 1994,
for a related approach).

The syntactic and semantic production rules for deriving the
feminine singular of a French adjective by suffixation with ``{\tt
e}'' are given, with some details omitted, in Figure \ref{morphderiv}.
In this case, nearly all features are shared between the inflected
word and the root, as is the logical form for the word (shown as
\verb!Adj! in the \verb!deriv! rule). The only differing feature is
that for gender, shown as the third argument of the \verb!@agr! macro,
which itself expands to a category.
\begin{figure*}
\begin{quote}
\begin{verbatim}
morph(adjp_adjp_fem,                      % rule (syntax)
  [adjp:[agr= @agr(3,sing,f) | Shared],   % mother category
   adjp:[agr= @agr(3,sing,m) | Shared],   % first daughter (category)
   e])                                    % second daughter (literal)
    :- Shared=[aform=Aform, ..., wh=n].   % shared syntactic features

deriv(adjp_adjp_fem, only                 % rule (semantics)
  [(Adj,adjp:Shared),                     % mother logical form and cat.
   (Adj,adjp:Shared),                     % first daughter
   (_,e)])                                % second daughter
    :- Shared=[anaIn=Ai, ..., subjval=Subj].  % shared semantic features
\end{verbatim}
\end{quote}
\caption{Syntactic and semantic morphological production rules}
\label{morphderiv}
\end{figure*}

Irregular forms, either complete words or affixable stems, are
specified by listing the morphological rules and terminal morphemes
from which the appropriate analyses may be constructed, for example:
\begin{verbatim}
   irreg(dit,[dire,'PRESENT_3s'],
             [v_v_affix-only]).
\end{verbatim}
Here, \verb!PRESENT_3s! is a pseudo-affix which has the same syntactic
and semantic information attached to it as (one sense of) the affix
``\verb!t!'', which is used to form some regular third person
singulars. However, the spelling rules make no reference to
\verb!PRESENT_3s!; it is simply a device allowing categories and logical
forms for irregular words to be built up using the same production
rules as for regular words.

\section{Compilation}

All rules and lexical entries in the CLE are compiled to a form that
allows normal Prolog unification to be used for category matching at
run time. The same compiled forms are used for analysis and
generation, but are indexed differently. Each feature for a major
category is assigned a unique position in the compiled Prolog term,
and features for which finite value sets have been specified are
compiled into vectors in a form that allows boolean expressions,
involving negation as well as conjunction and disjunction, to be
conjoined by unification (see Mellish, 1988; Alshawi, 1992, pp46--48).

The compilation of morphological information is motivated by the
nature of the task and of the languages to be handled. As discussed in
Section \ref{intro}, we expect the number of affix combinations to be
limited, but the lexicon is not necessarily known in advance.
Morphophonological interactions may be quite complex, and the purpose
of morphological processing is to derive syntactic and semantic
analyses from words and vice versa for the purpose of full NLP.
Reasonably quick compilation is required, and run-time speed need only
be moderate.

\subsection{Compiling Spelling Patterns}

Compilation of individual \verb!spell! rules is straightforward;
feature specifications are compiled to positional/boolean format,
characters and occurrences of character classes are also converted to
boolean vectors, and left contexts are reversed (cf Abrahamson, 1992)
for efficiency. However, although it would be possible to analyse
words directly with individually compiled rules (see Section
\ref{debugging} below), it can take an unacceptably long time to do
so, largely because of the wide range of choices of rule available at
each point and the need to check at each stage that obligatory rules
have not been broken. We therefore take the following approach.

First, all legal sequences of morphemes are produced by top-down
nondeterministic application of the production rules (Section
\ref{morphrules}), selecting affixes but keeping the root morpheme
unspecified because, as explained above, the lexicon is undetermined
at this stage.  For example, for English, the sequences
\verb!*+ed+ly! and \verb!un+*+ing!  are among those produced, the
asterisk representing the unspecified root.

Then, each sequence, together with any associated restrictions on
orthographic features, undergoes analysis by the compiled spelling
rules (Section \ref{spellrules}), with the surface sequence and the
root part of the lexical sequence initially uninstantiated.  Rules are
applied recursively and nondeterministically, somewhat in the style of
Abramson (1992), taking advantage of Prolog's unification mechanism
to instantiate the part of the surface string corresponding to affixes
and to place some spelling constraints on the start and/or end of the
surface and/or lexical forms of the root.

This process results in a set of {\it spelling patterns}, one for each
distinct application of the spelling rules to each affix sequence
suggested by the production rules. A spelling pattern consists of
partially specified surface and lexical root character sequences,
fully specified surface and lexical affix sequences, orthographic
feature constraints associated with the spelling rules and affixes
used, and a pair of syntactic category specifications derived from the
production rules used. One category is for the root form, and one for
the inflected form.

Spelling patterns are indexed according to the surface (for analysis)
and lexical (for generation) affix characters they involve. At run
time, an inflected word is analysed nondeterministically in several
stages, each of which may succeed any number of times including zero.
\begin{itemize}
\item stripping off possible (surface) affix characters in the word
and locating a spelling pattern that they index;
\item matching the remaining characters in the word against the
surface part of the spelling pattern, thereby, through shared
variables, instantiating the characters for the lexical part to
provide a possible root spelling;
\item checking any orthographic feature constraints on that root;
\item finding a lexical entry for the root, by any of a range of
mechanisms including lookup in the system's own lexicon, querying an
external lexical database, or attempting to guess an entry for an
undefined word; and
\item unifying the root lexical entry with the root category in the
spelling pattern, thereby, through variable sharing with the
other category in the pattern, creating a fully specified category
for the inflected form that can be used in parsing.
\end{itemize}
In generation, the process works in reverse, starting from
indexes on the lexical affix characters.

\subsection{Representing Lexical Roots}

Complications arise in spelling rule application from the fact that,
at compile time, neither the lexical nor the surface form of the root,
nor even its length, is known. It would be possible to hypothesize all
sensible lengths and compile separate spelling patterns for each.
However, this would lead to many times more patterns being produced
than are really necessary.

Lexical (and, after instantiation, surface) strings for the
unspecified roots are therefore represented in a more complex but less
redundant way: as a structure
\begin{quote}
$L_1$ $...$ $L_m$ {\tt v($L,R$)} $R_1$ ... $R_n$.
\end{quote}
Here the $L_i$'s are variables later instantiated to single characters
at the beginning of the root, and $L$ is a variable, which is later
instantiated to a list of characters, for its continuation. Similarly,
the $R_i$'s represent the end of the root, and $R$ is the continuation
(this time reversed) leftwards into the root from the $R_1$.  The {\tt
v($L,R$)} structure is always matched specially with a Kleene-star of
the {\tt default} spelling rule.  For full generality and minimal
redundancy, $L_m$ and $R_1$ are constrained not to match the default
rule, but the other $L_i$'s and $R_i$'s may. The values of $n$
required are those for which, for some spelling rule, there are $k$
characters in the target lexical string and $n-k$ from the beginning
of the right context up to (but not including) a boundary symbol.  The
lexical string of that rule may then match $R_1,...,R_k$, and its
right context match $R_{k+1},...,R_n,$\verb!+!$,...$. The required
values of $m$ may be calculated similarly with reference to the left
contexts of rules.\footnote{Alternations in the middle of a root, such
as umlaut, can be handled straightforwardly by altering the root/affix
pattern from $L_1 \ldots L_m$ {\tt v($L,R$)} $R_1 ... R_n$ to $L_1
\ldots L_m$ {\tt v($L,R$)} $M$ v($L',R'$) $R_1 ... R_n$, with $M$
forbidden to be the {\tt default} rule. This has not been necessary
for the descriptions developed so far, but its implementation is not
expected to lead to any great decrease in run-time performance,
because the non-determinism it induces in the lookup process is no
different in kind from that arising from alternations at root-affix
boundaries.}

During rule compilation, the spelling pattern that leads to the
run-time analysis of {\it ch\`{e}re} given above is derived from $m=0$
and $n=2$ and the specified rule sequence, with the variables $R_1$
$R_2$ matching as in Figure \ref{chere2}.
\begin{figure*}
\begin{center}
\begin{tabular}{|l|l|c|c|c|c|c|c|c|}\hline
Compile & Rule:    & \multicolumn{2}{c|}{$def.^*$} & {\it c.e\_\`{e}1} & $def.$
& $bdy.$ & $def.$ & $bdy.$
\\ \cline{2-9}
time: & Variable: & \multicolumn{2}{c|}{\tt v($L,R$)} & $R_1$ & $R_2$ &
\multicolumn{3}{c|}{...} \\ \hline \hline
Run & Surface: & c & h & \`{e} & r & & e & \\ \cline{2-9}
time: &Lexical: & c & h &  e    & r & + & e & + \\ \hline
\end{tabular}
\caption{Spelling pattern application to the analysis of {\it ch\`{e}re}}
\label{chere2}
\end{center}
\end{figure*}
\subsection{Applying Obligatory Rules}

In the absence of a lexical string for the root, the correct treatment
of obligatory rules is another problem for compilation. If an
obligatory rule specifies that lexical $X$ must be realised as surface
$Y$ when certain contextual and feature conditions hold, then a
partitioning where $X$ is realised as something other than $Y$ is only
allowed if one or more of those conditions is unsatisfied. Because of
the use of boolean vectors for both features and characters, it is
quite possible to constrain each partitioning by unifying it with the
complement of one of the conditions of each applicable obligatory
rule, thereby preventing that rule from applying. For English, with
its relatively simple inflectional spelling changes, this works well.
However, for other languages, including French, it leads to excessive
numbers of spelling patterns, because there are many obligatory rules
with non-trivial contexts and feature specifications.

For this reason, complement unification is not actually carried out at
compile time. Instead, the spelling patterns are augmented with the
fact that certain conditions on certain obligatory rules need to be
checked on certain parts of the partitioning when it is fully
instantiated. This slows down run-time performance a little but, as we
will see below, the speed is still quite acceptable.

\subsection{Timings}

The compilation process for the entire rule set takes just over a
minute for a fairly thorough description of French inflectional
morphology, running on a Sparcstation 10/41 (SPECint92=52.6). Run-time
speeds are quite adequate for full NLP, and reflect the fact that the
system is implemented in Prolog rather than (say) C and that full
syntactico-semantic analyses of sentences, rather than just morpheme
sequences or acceptability judgments, are produced.

Analysis of French words using this rule set and only an in-core
lexicon averages around 50 words per second, with a mean of 11
spelling analyses per word leading to a mean of 1.6 morphological
analyses (the reduction being because many of the roots suggested by
spelling analysis do not exist or cannot combine with the affixes
produced). If results are cached, subsequent attempts to analyse the
same word are around 40 times faster still. Generation is also quite
acceptably fast, running at around 100 words per second; it is
slightly faster than analysis because only one spelling, rather than
all possible analyses, is sought from each call. Because of the
separation between lexical and morphological representations, these
timings are essentially unaffected by in-core lexicon size, as full
advantage is taken of Prolog's built-in indexing.

Development times are at least as important as computation times. A
rule set embodying a quite comprehensive treatment of French
inflectional morphology was developed in about one person month. The
English spelling rule set was adapted from Ritchie {\it et al} (1992)
in only a day or two. A Polish rule set is also under development,
and Swedish is planned for the near future.

\section{Some Examples}

To clarify further the use of the formalism and the operation of the
mechanisms, we now examine several further examples.

\subsection{Multiple-letter spelling changes}
\label{eauelle}

Some obligatory spelling changes in French involve more than one
letter. For example, masculine adjectives and nouns ending in {\it
eau} have feminine counterparts ending in {\it elle}: {\it beau}
(``nice'') becomes {\it belle}, {\it chameau} (``camel'') becomes {\it
chamelle}. The final {\it e} is a feminizing affix and can be seen as
inducing the obligatory spelling change {\it au $\rightarrow$ ll}.
However, although the obvious spelling rule,
\begin{center}
\verb!spell(change_au_ll, "|ll|"! $\leftrightarrow$ \verb!"|au|+e")!,
\end{center}
allows this change, it does not rule out the incorrect realization of
\verb!beau+e! as {\it *beaue}, shown in Figure \ref{beaue},
\begin{figure*}
\begin{center}
\begin{tabular}{|l|c|c|c|c|c|c|c|}\hline
Surface: & b & e & a & u & & e & \\
Lexical: & b & e & a  & u & + & e & + \\ \hline
Rule:    & $def.$ & $def.$ & $def.$ & $def.$ & $bdy.$ & $def.$ & $bdy.$
\\ \hline
\end{tabular}
\caption{Incorrect partitioning for {\tt beau+e+}}
\label{beaue}
\end{center}
\end{figure*}
because it only affects partitionings where the \verb!au! at the
lexical level forms a {\it single} partition, rather than one for
\verb!a! and one for \verb!u!. Instead, the following pair of rules,
in which the lexical targets have only one character each, achieve the
desired effect:
\pagebreak
\begin{center}
\verb!spell(change_au_ll1, "|l|"! $\leftrightarrow$ \verb!"|a|u+e")! \\
\verb!spell(change_au_ll2, "|l|"! $\leftrightarrow$ \verb!"a|u|+e")!
\end{center}
Here, \verb!change_au_ll1! rules out {\tt a:a} partition
in Figure \ref{beaue}, and \verb!change_au_ll2! rules out the
{\tt u:u} one.

It is not necessary for the {\it surface} target to contain exactly
one character for the blocking effect to apply, because the semantics
of obligatoriness is that the {\it lexical} target and all contexts,
taken together, make the specified {\it surface} target (of whatever
length) obligatory for that partition. The reverse constraint, on the
lexical target, does not apply.

\subsection{Using features to control rule application}

Features can be used to control the application of rules to particular
lexical items where the applicability cannot be deduced from spellings
alone. For example, Polish nouns with stems whose final syllable has
vowel {\it \'{o}} normally have inflected forms in which the accent is
dropped. Thus in the nominative plural, {\it kr\'{o}j} (``style'')
becomes {\it kroje}, {\it b\'{o}r} (``forest'') becomes {\it bory},
{\it b\'{o}j} (``combat'') becomes {\it boje}. However, there are
exceptions, such as {\it zb\'{o}j} (``bandit'') becoming {\it
zb\'{o}je}. Similarly, some French verbs whose infinitives end in {\it
-eler} take a grave accent on the first {\it e} in the third person
singular future ({\it modeler}, ``model'', becomes {\it
mod\`{e}lera}), while others double the {\it l} instead (e.g.\ {\it
appeler}, ``call'', becomes {\it appellera}).

These phenomena can be handled by providing an obligatory rule for the
case whether the letter changes, but constraining the applicability
of the rule with a feature and making the feature clash with that
for roots where the change does not occur. In the Polish case:

\vspace{2mm}
\noindent
\hspace*{0.5cm}{\tt spell(change\_\'{o}\_o, "|o|" $\leftrightarrow$
"|\'{o}|1+2", \newline
\hspace*{1.5cm}[1/c, 2/v], [chngo=y]). \newline
\ \newline
\hspace*{0.5cm}orth(zb\'{o}j, [chngo=n]).} \newline

\noindent
Then the partitionings given in Figure \ref{zboj} will be the only
possible ones. For {\it b\'{o}j}, the {\tt change\_\'{o}\_o} rule
must apply, because the \verb!chngo! feature for {\it b\'{o}j} is
unspecified and therefore can take any value; for {\it zb\'{o}j},
however, the rule is prevented from applying by the feature clash,
and so the default rule is the only one that can apply.
\begin{figure*}
\begin{center}
\begin{tabular}{|l|c|c|c|c|c|c|}\hline
Surface: \hspace*{8mm}  & b & o & j &  & e & \\
Lexical: & b & \'{o} & j & + & e & + \\ \hline
Rule:    & $def.$ & {\it c\_\'{o}\_o.} & $def.$
& $bdy.$ & $def.$ & $bdy.$ \\ \hline
\end{tabular}

\vspace{2mm}
\begin{tabular}{|l|c|c|c|c|c|c|c|}\hline
Surface: & z & b & \'{o} & j &  & e & \\
Lexical: & z & b & \'{o} & j & + & e & + \\ \hline
Rule:    & $def.$ & $def.$ & $def.$ & $def.$ & $bdy.$ & $def.$ & $bdy.$
\\ \hline
\end{tabular}
\caption{Feature-dependent dropping of accent}
\label{zboj}
\end{center}
\end{figure*}

\section{Debugging the Rules}
\label{debugging}

The debugging tools help in checking the operation of the spelling
rules, either (1) in conjunction with other constraints or (2) on
their own.

For case (1), the user may ask to see all inflections of a root
licensed by the spelling rules, production rules, and lexicon; for
{\it cher}, the output is
\begin{quote}
\verb![cher,e]: adjp -> !{\tt ch\`ere} \\
\verb![cher,e,s]: adjp -> !{\tt ch\`eres} \\
\verb![cher,s]: adjp -> chers!
\end{quote}
meaning that when {\it cher} is an \verb!adjp! (adjective) it may
combine with the suffixes listed to produce the inflected forms shown.
This is useful in checking over- and undergeneration. It is also
possible to view the spelling patterns and production rule tree used
to produce a form; for {\it ch\`{e}re}, the trace (slightly simplified
here) is as in figure \ref{chertrace}.
\begin{figure*}
\begin{quote}
{\tt
"ch\`{e}re" has root "cher" with pattern 194 and tree 17. \\
\  \\
Pattern 194: \\
\begin{quote}
\verb!"___!\`{e}\verb!{clmnprstv=A}e" <-> "___e{clmnprstv=A}+e+"! \\
\verb!=> tree 17 and 18 if [doublec=n]! \\
\verb!Uses: default* change_e_!\`{e}\verb!1 default boundary default boundary!
\\
\end{quote}
Tree 17: \\
\begin{verbatim}
   Both = adjp:[dmodified=n,headfinal=y,mhdfl=y,synmorpha=1,wh=n]
   Root = adjp:[agr=agr:[gender=m]]
   Infl = adjp:[agr=agr:[gender=f]]
   Tree = adjp_adjp_fem=>[*,e]
\end{verbatim}
}
\caption{Debugger trace of derivation of {\it ch\`{e}re}}
\label{chertrace}
\end{quote}
\end{figure*}
The spelling pattern 194 referred to here is the one depicted in a
different form in Figure \ref{chere2}. The notation
\verb!{clmnprstv=A}! denotes a set of possible consonants represented
by the variable \verb!A!, which also occurs on the right hand side of
the rule, indicating that the same selection must be made for both
occurrences.  Production rule tree 17 is that for a single application
of the rule \verb!adjp_adjp_fem!, which describes the feminine form of
the an adjective, where the root is taken to be the masculine form.
The \verb!Root! and \verb!Infl!  lines show the features that differ
between the root and inflected forms, while the \verb!Both! line shows
those that they share.  Tree 18, which is also pointed to by the
spelling pattern, describes the feminine forms of nouns analogously.

For case (2), the spelling rules may be applied directly, just as in
rule compilation, to a specified surface or lexical character
sequence, as if no lexical or morphotactic constraints existed.
Feature constraints, and cases where the rules will not apply if those
constraints are broken, are shown.  For the lexical sequence
\verb!cher+e+!, for example, the output is as follows.
\begin{quote}
\verb!Surface: "ch!{\tt \`e}\verb!re" <->! \\
\verb!Lexical: "cher". Suffix: "e"! \\
\verb!c :: c <- default! \\
\verb!h :: h <- default! \\
{\tt \`e}\verb! :: e <- change_e_!{\tt \`e1} \\
\verb!r :: r <- default!\\
\hspace*{0mm}\verb!  :: + <- boundary!\\
\verb!Category: orth:[cdouble=n]!\\
\verb!e :: e <- default!\\
\hspace*{0mm}\verb!  :: + <- boundary!\\
\end{quote}
\begin{quote}
\verb!Surface: "chere" <->! \\
\verb!Lexical: "cher". Suffix: "e"!\\
\verb!c :: c <- default! \\
\verb!h :: h <- default! \\
\verb!e :: e <- default  (breaks "change_e_!{\tt \`e}\verb!1")! \\
\verb!r :: r <- default!\\
\hspace*{0mm}\verb!  :: + <- boundary!\\
\verb!e :: e <- default!\\
\hspace*{0mm}\verb!  :: + <- boundary!\\
\end{quote}
This indicates to the user that if {\it cher} is given a lexical entry
consistent with the constraint \verb!cdouble=n!, then only the first
analysis will be valid; otherwise, only the second will be.

\section{Conclusions and Further Work}

The rule formalism and compiler described here work well for European
languages with reasonably complex orthographic changes but a limited
range of possible affix combinations. Development, compilation and
run-time efficiency are quite acceptable, and the use of rules
containing complex feature-augmented categories allows morphotactic
behaviours and non-segmental spelling constraints to be specified in a
way that is perspicuous to linguists, leading to rapid development of
descriptions adequate for full NLP.

The kinds of non-linear effects common in Semitic languages, where
vowel and consonant patterns are interpolated in words (Kay, 1987;
Kiraz, 1994) could be treated efficiently by the mechanisms described
here if it proved possible to define a representation that allowed the
parts of an inflected word corresponding to the root to be separated
fairly cleanly from the parts expressing the inflection. The latter
could then be used by a modified version of the current system as the
basis for efficient lookup of spelling patterns which, as in the
current system, would allow possible lexical roots to be calculated.

Agglutinative languages could be handled efficiently by the current
mechanism if specifications were provided for the affix combinations
that were likely to occur at all often in real texts. A backup
mechanism could then be provided which attempted a slower, but more
complete, direct application of the rules for the rarer cases.

The interaction of morphological analysis with spelling correction
(Carter, 1992; Oflazer, 1994; Bowden, 1995) is another possibly fruitful
area of work. Once the root spelling patterns and the affix
combinations pointing to them have been created, analysis essentially
reduces to an instance of affix-stripping, which would be
amenable to exactly the technique outlined by Carter (1992). As in
that work, a discrimination net of root forms would be required;
however, this could be augmented independently of spelling pattern
creation, so that the flexibility resulting from not composing the
lexicon with the spelling rules would not be lost.

\section*{Acknowledgments}

I am grateful to Manny Rayner and anonymous European ACL referees for
commenting on earlier versions of this paper, and to Pierrette
Bouillion and Ma\l gorzata Sty\'{s} for comments and also for
providing me with their analyses of the French and Polish examples
respectively.

This research was partly funded by the Defence Research Agency,
Malvern, UK, under Strategic Research Project M2YBT44X.

\section*{References}

\newenvironment{reverseindent}%
{\begin{list}{}{\setlength{\labelsep}{0in}
                \setlength{\labelwidth}{0in}
                \setlength{\itemindent}{-\leftmargin}}}%
{\end{list}}

\begin{reverseindent}

\item
Abramson, H., (1992). ``A Logic Programming View of Relational
Morphology''. {\it Proceedings of COLING-92}, 850--854.

\item
Alshawi, H.\ (1992). {\it The Core Language Engine} (ed). MIT Press.

\item
Alshawi, H., D.J.\ Arnold, R.\ Backofen, D.M.\ Carter, J.\ Lindop, K.\
Netter, S.G.\ Pulman, J.\ Tsujii, and H.\ Uszkoreit (1991). {\it Eurotra
ET6/1: Rule Formalism and Virtual Machine Design Study}. Commission of
the European Communities, Luxembourg.

\item
Bowden, T.\ (1995) ``Cooperative Error Handling and Shallow
Processing'', these proceedings.

\item
Carter, D.M.\ (1992). ``Lattice-based Word Identification in CLARE''.
{\it Proceedings of ACL-92}.

\item
Kaplan, R., and M.\ Kay (1994). ``Regular Models of Phonological Rule
Systems'', {\it Computational Linguistics}, 20:3, 331--378.

\item
Kay, M.\ (1987). ``Non-concatenative Finite-State Morphology''. {\it
Proceedings of EACL-87}.

\item
Karttunen, L., R.M.\ Kaplan, and A.\ Zaenen (1992). ``Two-level
Morphology with Composition''.  {\it Proceedings of COLING-92},
141--148.

\item
Kiraz, G. (1994). ``Multi-tape Two-level Morphology''.  {\it
Proceedings of COLING-94}, 180--186.

\item
Koskenniemi, K.~(1983). {\it Two-level morphology: a general
computational model for word-form recognition and production}.
University of Helsinki, Department of General Linguistics,
Publications, No.~11.

\item
Kwon, H-C., and L.\ Karttunen (1994). ``Incremental Construction of
a Lexical Transducer for Korean''.   {\it Proceedings of COLING-94},
1262--1266.

\item
Mellish, C.~S.\ (1988). ``Implementing Systemic Classification by
Unification''.  {\it Computational Linguistics} 14:40--51.

\item
Oflazer, K.\ (1993). ``Two-level Description of Turkish Morphology''.
{\it Proceedings of European ACL-93}.
\item

Oflazer, K.\ (1994). {\it Spelling Correction in Agglutinative
Languages}. Article \verb!9410004! in \newline \verb!cmp-lg@xxx.lanl.gov!
archive.

\item
Ritchie, G., G.J.\ Russell, A.W.\ Black and S.G.\ Pulman (1992).
{\it Computational Morphology}. MIT Press.

\item
Ruessink, H.\ (1989). {\it Two Level Formalisms}. Utrecht Working
Papers in NLP, no.\ 5.

\item
Trost, H.\ (1990). ``The Application of Two-level Morphology to
Non-Concatenative German Morphology''. {\it Proceedings of COLING-90},
371--376.

\item
Trost, H.\ (1991). ``X2MORF: A Morphological Component Based on
Augmented Two-level Morphology''. {\it Proceedings of IJCAI-91},
1024--1030.

\item
Trost, H., and J.\ Matiasek (1994). ``Morphology with a
Null-Interface'', {\it Proceedings of COLING-94}.

\end{reverseindent}

\end{document}